\documentclass[11pt,a4paper]{article}
\usepackage[hyperref]{emnlp2020}
\usepackage{times}
\usepackage{latexsym}
\usepackage{todonotes}

\usepackage{amsmath,amsfonts,bm}

\def\eqref#1{equation~\ref{#1}}

\def\1{\bm{1}}

\DeclareMathAlphabet{\mathsfit}{\encodingdefault}{\sfdefault}{m}{sl}
\SetMathAlphabet{\mathsfit}{bold}{\encodingdefault}{\sfdefault}{bx}{n}

\usepackage{multirow}
\usepackage{tikz,pgfplots}
\usepackage{booktabs} 

\usepackage{microtype}

\aclfinalcopy 

\newcommand{\change}[1]{%
{#1}
}

\title{Vector-Vector-Matrix Architecture: A Novel Hardware-Aware Framework for Low-Latency Inference in NLP Applications}

\author{Matthew Khoury\textsuperscript{*,1}, \  Rumen Dangovski\textsuperscript{*,\dag,2},  \ Longwu Ou\textsuperscript{1}, \\ \textbf{Preslav Nakov\textsuperscript{3}, \ Yichen Shen\textsuperscript{2}, \ Li Jing\textsuperscript{\dag,1}} \\ 
 (*) equal contribution, (\dag) work done at Lightelligence, Inc. \\
 \textsuperscript{1}Lightelligence, Inc. \\ \texttt{\{matthew.khoury, longwu.ou, yichen, li\}@lightelligence.ai} \\
 \textsuperscript{2}Massachusetts Institute of Technology \\ \texttt{rumenrd@mit.edu} \\ 
 \textsuperscript{3}Qatar Computing Research Institute, HBKU \\ \texttt{pnakov@hbku.edu.qa}

 }

\date{}

\begin{document}
\maketitle
\begin{abstract}
Deep neural networks have become the standard approach to building reliable Natural Language Processing (NLP) applications, ranging from Neural Machine Translation (NMT) to dialogue systems. However, improving accuracy by increasing the model size requires a large number of hardware computations, which can slow down NLP applications significantly at inference time. To address this issue, we propose a novel vector-vector-matrix architecture (VVMA), which greatly reduces the latency at inference time for NMT. This architecture takes advantage of specialized hardware that has low-latency vector-vector operations and higher-latency vector-matrix operations. It also reduces the number of parameters and FLOPs for virtually all models that rely on efficient matrix multipliers without significantly impacting accuracy. We present empirical results suggesting that our framework can reduce the latency of sequence-to-sequence and Transformer models used for NMT by a factor of four. Finally, we show evidence suggesting that our VVMA extends to other domains, and we discuss novel hardware for its efficient use.

\end{abstract}

\begin{figure}[tbh]
\begin{center}
\includegraphics[width=0.48 \textwidth]{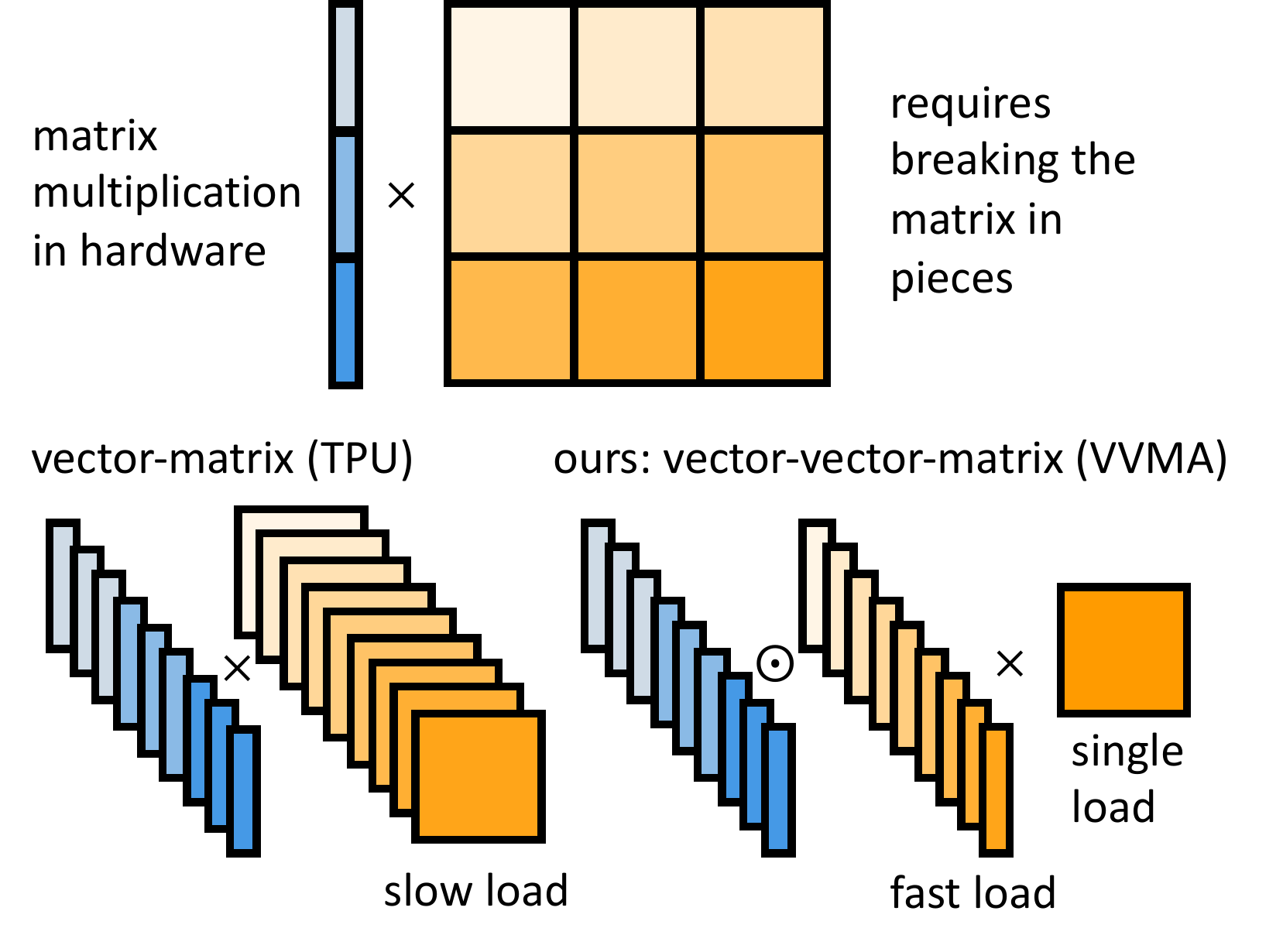}
\caption{TPU vs. VVMA. Top: to multiply a vector by a matrix, the hardware tiles up the matrix. Bottom left: the TPU loads each piece. Bottom right: the VVMA loads a single piece (for broadcasting) and adds diagonals for element-wise multiplication, which is faster.}\label{figure:1}
\end{center}
\end{figure}

\section{Introduction}

Artificial neural networks have become increasingly popular over the last decade as they excel in tasks such as object detection and speech recognition {\cite{deep_learning}}, which are becoming more commonplace with the use of self-driving cars and virtual assistants.
The rapid development of deep neural networks has also made them the dominant approach for natural language processing (NLP) applications, ranging from neural machine translation (NMT) \cite{Bahdanau2014NeuralMT, Klein2017OpenNMTOT, 45610} and text summarization \cite{Rush2015ANA, Nallapati2016AbstractiveTS, Liu2018GeneratingWB} to virtual assistants such as Apple Siri, Amazon Alexa, and Google Home.

Unfortunately, neural networks are slow for training, inference and use due to their vast computational complexity. Several approaches have been proposed to address these issues including (\emph{a})~quantization and pruning,
(\emph{b})~efficient models with less computational demand,
and (\emph{c})~specialized hardware accelerators~\cite{ml_overview}.
While direction (\emph{a}) has been well-studied~\cite{optimal_brain_damage, pruning_and_quantization, sparse_lstm, quantization_survey, quinn2018}, and can be considered complementary to (\emph{b,c}),
optimizing the combination of (\emph{b}) and (\emph{c}) 
has not been considered, to the best of our knowledge. Thus, here we propose a novel {\emph{vector-vector-matrix architecture}} (VVMA) that compresses neural networks, while optimizing for hardware performance at inference time. Therefore, we optimize (\emph{b}) and (\emph{c}), without conflicting with (\emph{a}), i.e.,~using quantization and pruning can potentially further boost the efficiency of our framework.  Figure~\ref{figure:1} illustrates this VVMA in contrast to a traditional vector-matrix architecture.

Moreover, the inherently sequential nature of many NLP tasks can increase the latency at inference time. Constrained by their memory bandwidth and footprint, modern accelerators rely on large batch sizes to avoid under-utilization. However, it is not always possible to increase the batch size if conclusions have to be inferred quickly, e.g.,~for real-time inference. For example, the matrix multiply unit of state-of-the-art accelerators, such as Google's Tensor Processing Unit (TPU), will ``stall'' when translating a single sentence, thus increasing the overall latency \cite{tpu}. 

Our architecture can improve the TPU and other AI accelerators for \textit{small-batch inference}. Thus, unlike other methods for compressing neural networks, the VVMA is designed to take advantage of the dataflow and the architecture of certain kinds of hardware accelerators such as the TPU.

Our contributions are as follows: 

\begin{itemize}
    \item We tailor an efficient model to state-of-the-art hardware accelerators.
    \item We provide an efficient vector-vector-matrix architecture (VVMA) framework for inference with small batch sizes.
    \item We use VVMAs to speed up inference in the computationally expensive Neural Machine Translation (NMT) task by a factor of four without losing much in terms of quality.
\item We highlight promising applications of the VVMA in other deep learning domains and novel Artificial Intelligence~(AI) accelerators.
\end{itemize}

The rest of this paper is organized as follows: In Section~\ref{background}, we elaborate on directions (\emph{b}) and (\emph{c}), and we relate them to VVMAs. In Section~\ref{architecture}, we motivate VVMAs as a faster improvement of the TPU's architecture and dataflow at inference time, and we then analyze our framework in its universality, including tips for efficient implementation of VVMAs. As a proof of concept, in Section~\ref{results} we demonstrate empirical inference speed-ups for NMT using Seq2seq-LSTM and Transformer models, which are both notorious for their computational complexity. We also show ablation studies and extensions to other tasks. In Section~\ref{discussion}, we explore novel accelerators that can benefit from VVMAs. Finally, we offer concluding remarks in Section~\ref{conclusion}, and we point to possible directions for future work.

\section{Background}
\label{background}

Here, we look at efficient models from the software and the hardware side, and we discuss the advantages of merging them in a \emph{co-design} manner. We further discuss the importance of wall-clock speed versus floating point operations and why from this perspective our weight sharing matrices will decrease inference rather than training time.

\subsection{Efficient Models from the Software Side for Training and Inference}

Efficient model architectures can decrease the complexity of neural networks. 
Some techniques to achieve this are described in \cite{hashing,shuffle,group_rnn}.

{\citet{shuffle}} added a new type of layer, a \emph{channel shuffle layer}, to neural networks that use group convolution. By shuffling the data between layers, they reduced the number of parameters in the other layers while retaining similar accuracy.
{\citet{group_rnn}} used a technique similar to group convolution, but applied it to recurrent neural networks. They used shuffling operations with a group recurrent neural network and showed improvements for NMT and text summarization.
 
{\citet{hashing}} compressed a weight matrix into a learned vector of weights. They used a hash function to map entries in the weight matrix to elements in the vector. As a result, many matrix entries share a single weight in the vector.

As Transformers are becoming the standard building block for NLP tasks, there is a growing effort to make them efficient, since their inference time scales as $O(N^2)$, where $N$ is the number of input tokens.
\citet{Child2019GeneratingLS} proposed Sparse Transformers with $O(N\sqrt{N})$ complexity.
Likewise,~\citet{Sukhbaatar2019AdaptiveAS} developed Adaptive Attention Span and~\citet{Kitaev2020ReformerTE} proposed Reformer using locality-sensitive hashing, and achieved $O(N\log N)$ complexity.
See \cite{ganesh2020compressing} for a broader overview.

In a similar fashion, our VVMA is an efficient model because it reduces the computational complexity at inference time without much decrease in performance. However, unlike the above models, VVMAs focus on the low levels of execution: the VVMA is an architecture that speeds up matrix multiplications. Thus, it is an efficient model that relates to hardware accelerators directly and it is universal, as matrix multiplication is the dominant computational factor for neural network inference.

\subsection{Efficient Models from the Hardware Side}

As we have mentioned above, successful NLP applications have been based on a variety of neural network models: recurrent and convolutional neural networks, memory-augmented networks, attention mechanism, Transformers, etc. These models were designed to solve numerous tasks ranging from language modeling and named entity recognition to NMT and other sequence modeling and sequence generation tasks. Most of the computation in such models is matrix multiplication both at inference and at training time, which is expensive. Therefore, specialized hardware accelerators for neural networks have been designed, focusing on making matrix multiplication efficient.

Note that the above techniques assume general hardware, i.e.,~they do not utilize the specific dataflow or architecture of an AI accelerator to improve efficiency.
Yet, several such accelerators have been developed recently, e.g.,~the Horizon Robotics Brain Processing Unit, Graphcore Intelligence Processing Unit, NVIDIA Tensor Core, and Google Tensor Processing Unit (TPU). 

A \textit{matrix-matrix} architecture is a hardware unit that takes two matrices and multiplies them, e.g.,~NVIDIA Tensor Core. A \textit{vector-matrix} architecture such as Google's TPU multiplies a vector and a matrix. As shown in Figure~{\ref{figure:1}}, the VVMA \textit{vector-vector-matrix} architecture takes two vectors and a matrix, and it multiplies element-wise the first vector by the second vector, and then multiplies the resulting vector by the matrix.

Furthermore, VVMAs are optimized for certain AI accelerators, such as the TPU architecture. We specifically take advantage of the dataflow of the matrix multiply unit in the TPU, which is described in \cite{tpu}. This matrix multiply unit allows to re-use weights for multiple batches of data, while also using a systolic loop to perform matrix multiplication extremely fast. Therefore, we reduce the computational complexity of the ``matrix'' component in the TPU's vector-matrix unit, but we also maintain representational accuracy by inserting an extra ``vector'' part to get the vector-vector-matrix unit. By switching to this unit, we introduce a trade-off by increasing the efficiency of the model while decreasing its flexibility and generalization power. Likewise, we expect to have comparable accuracy to other compression techniques while also providing even faster performance at inference time.

\subsection{Trade-Off between Flexibility and Efficiency at Inference Time}

While every neural network requires a certain budget of floating point operations for a target computation, how fast such computations are in practice depends not on the size of this budget but rather on the number of wall clocks needed in order to cover all floating point operations. Thus, it is important to combine the software and the hardware advances in a co-design manner to optimize an efficient model for the correct metric: wall clocks.

Designed to optimize for the number of wall clocks, our \change{VVMA} introduces an extra vector component that maintains accuracy, but increases the computational complexity. We achieve this in part by optimizing our VVMA to specifically take advantage of the TPU architecture and dataflow. This creates a trade-off between flexibility and efficiency, e.g.,~the more we reuse weights, the more we have to compensate for the model accuracy.  

Neural networks that are specifically designed to work in conjunction with certain AI accelerators will encounter a similar trade-off. That is, the more a neural network is tuned for efficiency, the less flexibility for change the model will have {\cite{more_pruning}}. Nonetheless, we find regimes that suppress this trade-off and yield faster neural networks inference with VVMA. Thus, we believe that our VVMAs provide enough flexibility to be useful in a variety of existing neural architectures.

Training is the process of using (large) datasets to learn specific weights in neural networks. This process is usually very computationally expensive and can take days or months to complete. Once a neural network has finished training, the set of weights that were learned through the training process can remain fixed while making predictions. This process of using a fixed set of weights to make predictions with a neural network is called \emph{inference} {\cite{ml_overview}}.
Training can be done faster when parallelizing the process and increasing the amount of data fed into the network at a given time. This roughly translates to increasing the throughput of the training process. However, when performing inference on a single data point, the latency of making predictions seems to dominate the run-time {\cite{tpu}}.
The VVMA we propose can be used specifically to decrease the latency of a neural network. Likewise, we expect this technique to be used to decrease inference time rather than to decrease training time.

\section{Architecture}
\label{architecture}

In this section, we present our approach to constructing a VVMA, including implementation details that are necessary to use VVMAs in practice. 

\subsection{Motivation}
Google announced their first application-specific AI accelerator called the \emph{Tensor Processing Unit} (TPU) in 2016. 
As described by {\citet{tpu}}, the TPU uses a systolic loop to perform matrix multiplications \cite{tpu}, which are the most demanding computations in deep neural networks. 
Let $W$ be an $n \times n$ weight matrix and $x$ be an $n$-dimensional input vector. In order to perform $Wx$ on the TPU, we must first break up $W$ and $x$ into $k \times k$ sections, where $k \times k$ is the size of the matrix multiply unit:
\begin{equation}
  Wx =
  \begin{bmatrix}
    W_{1, 1} & W_{1, 2} & \cdots \\
    W_{2, 1} & W_{2, 2} & \cdots \\
    \vdots & \vdots & \ddots 
  \end{bmatrix}
  \begin{bmatrix}
    x_1 \\
    x_2 \\
    \vdots
  \end{bmatrix}.
\end{equation}

Here, $W_{i, j}$ is a $k \times k$ block of $W$, and $x_j$ is a $k$-dimensional block of $x$. Likewise, the TPU must load each block $W_{i, j}$ onto the matrix multiply unit before multiplying it by $x_{j}$. Loading a $k \times k$ block takes $O(k)$ clocks on the TPU. After loading a block $W_{i,j}$ onto the TPU, it takes $O(2k + t)$ clocks to multiply $t$ $k$-dimensional vectors $x_j$ by the matrix $W_{i, j}$. So, the total number of clocks to multiply $t$ $n$-dimensional vectors $x$ by $W$ is 

\begin{equation}\label{original_clocks}
    O\left(\frac{n^2}{k^2} (k + 2k + t) \right).
\end{equation}

\begin{figure}[t]
  \centering 
  \tikz \node [scale=1, inner sep=0] {
  \begin{tikzpicture}
    \node[minimum size = 1cm, circle, draw, fill=green!10, ultra thick]
    at (0, 0) (x1) {$x_j$};
    \node[minimum height=1.5cm, minimum width=1.5cm, rectangle, rounded corners, draw, fill=cyan!10, ultra thick]
    at (2, 1) (w1) {$W_{i, j}$};
    \node[minimum height=1cm, minimum width=2.5cm, rectangle, rounded corners=15pt, draw, fill=yellow!10, ultra thick]
    at (5.5, 1) (out1) {$W_{i, j} x_j$}; 
    \node[minimum height=1.5cm, minimum width=.75cm, rectangle, draw, rounded corners, fill=red!10, ultra thick]
    at (1.25, -1) (v2) {$v_{i,j}$}; 
    \node[minimum height=1.5cm, minimum width=1.5cm, rectangle, draw, rounded corners, fill=cyan!10, ultra thick]
    at (3, -1) (w2) {$M$};
    \node[minimum height=1cm, minimum width=2.5cm, rectangle, rounded corners=15, draw, fill=yellow!10, ultra thick]
    at (5.5, -1) (out2) {$M (v_{i,j} \odot x_j)$};
    \draw [ultra thick, ->] (x1.north) to [out=90, in=180] (w1);
    \draw [ultra thick, ->] (w1) to (out1);
    \draw [ultra thick, ->] (v2) to (w2);
    \draw [ultra thick, ->] (w2) to (out2);
    \draw [ultra thick, ->] (x1.south) to [out=270, in=180] (v2); 
  \end{tikzpicture}
  };
  \caption{Illustration of how we can save time by sharing a weight matrix $M$. The top path shows the traditional dataflow, where each $W_{i, j}$ must be loaded onto the matrix multiply unit. The bottom path shows our approach, where $M$ is loaded onto the matrix multiply unit only once. We then add a vector-vector operation $v_{i,j}  \odot x_j$ before doing the matrix multiplication, where $\odot$ denotes element-wise multiplication.}\label{figure:2}
\end{figure}
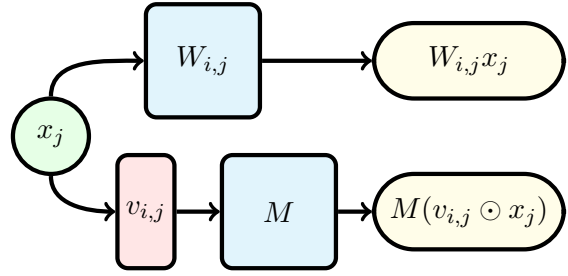

Note the large latency for single-batch inference, i.e.,~for $t=1$.
In order to decrease it, we tweak the weight matrix $W$, so that we only have to load a single $k \times k$ block $M$ onto the matrix multiply unit. We then perform vector operations to each $x_{j}$ 
in order to make up for the extra 
parameters that are lost by re-using the same $k \times k$ matrix $M$. 
Figure~{\ref{figure:2}} shows an illustration of this process. 

With this new procedure, the total number of clocks to multiply $t$ $n$-dimensional vectors by the larger matrix is given by
\begin{equation}\label{weight_sharing_clocks}
     O\left(k + 2k + \frac{n^2 t}{k^2}\right).
\end{equation}

We can see that this new procedure significantly decreases the total number of clocks for single-batch inference with $t=1$.

\subsection{Vector-Vector-Matrix Architecture}

We construct the VVMA as follows. 
Let $W$ be a large $n \times n$ weight matrix and let $M$ be a smaller $k \times k$ weight matrix. First, we tile $M$ into a larger matrix, so that its size is greater than or equal to the weight matrix $W$. 
Then, we multiply each copy of $M$ by a unique diagonal matrix. 
Mathematically, we replace $W$ with a structured matrix as shown below:

\begin{equation}\label{shared_matrix}
  \begin{bmatrix}
    M D_{1, 1} & M D_{1, 2} & \cdots \\
    M D_{2, 1} & M D_{2, 2} & \cdots \\
    \vdots & \vdots & \ddots 
  \end{bmatrix},
\end{equation} 

\noindent where $M$ is a shared $k \times k$ weight matrix and $D_{i, j}$ is a diagonal $k \times k$ weight matrix. 

We use the diagonal matrices $D_{i, j}$ in order to introduce variation to each of the copies of $M$. We found that this is necessary for a VVMA to be able to effectively replace the original matrix $W$. 
Each of the entries in the matrix $M$ is shared in multiple blocks of the new matrix, thus decreasing the total number of parameters compared to the original weight matrix $W$. Moreover, each of the entries of $M$ as well as the entries in each diagonal matrix $D_{i, j}$ are learned as part of the training process.

Even though each entry $D_{i, j}$ is mathematically represented as a matrix in Equation {\ref{shared_matrix}}, we can also see it as a $k$-dimensional vector $v_{i, j}$. We can then perform the matrix multiplication $D_{i, j} x$ as an element-wise multiplication $v_{i,j} \odot x_j$, as shown in Figure~{\ref{figure:2}}.

\begin{table*}[tb]
  \small
  \centering
  \begin{tabular}{lllllll} \toprule		
    \bf Task & \bf Model & \bf Architecture & \bf \# Params & \bf BLEU & \bf \# Clocks & \bf FLOPs \\
    \midrule
    German-English & Seq2Seq-LSTM & Original & 
    210.9M & 22.42 & 322.1M & 421.7M
    \\
    & & VVMA & 
    115.4M & 21.53 & 98.3M & 230.9M
    \\
    & Transformer & Original & 
    61.4M & 29.66 & 145.2M & 122.6M
    \\
    & & VVMA & 
    18.8M & 23.32 & 42.2M & 37.5M
    \\ \midrule 
    English-German & Seq2Seq-LSTM & Original & 
    210.9M & 20.70 & 322.1M & 421.7M
    \\
    & & VVMA & 
    115.4M & 18.90 & 98.3M & 230.9M
    \\
    & Transformer & Original & 
    61.4M & 24.57 & 145.2M & 122.6M
    \\
    & & VVMA & 
    18.8M & 18.99 & 42.2M & 37.5M
    \\ \midrule 
    Vietnamese-English & Seq2Seq-LSTM & Original & 
    32.3M & 22.42 & 46.3M & 64.6M
    \\
    & & VVMA & 
    21.9M & 20.86 & 21.9M & 43.8M
    \\ \midrule 
    English-Vietnamese & Seq2Seq-LSTM & Original & 
    27.5M & 25.34 & 34.8M & 54.9M
    \\
    & & VVMA & 
    17.1M & 24.42 & 10.3M & 34.1M
    \\ \bottomrule 
  \end{tabular}
  \caption{Comparing the original Seq2seq-LSTM and Transformer models to such with VVMAs. Shown are the number of parameters, the BLEU score, and the estimated number of clock cycles and floating point operations.}\label{table:all:nmt}
\end{table*}

\subsection{Implementation Details}

In order to implement \eqref{shared_matrix} as a trainable matrix, we found that it was inefficient to actually construct the entire matrix representation. Instead, it was better to take advantage of broadcasting, which allows us to element-wise multiply tensors of different shapes. Likewise, we use broadcasting to multiply the input vector $x$ by a larger diagonal tensor $D$. We then perform a matrix multiplication with the broadcasted vector and the matrix $M$. Thus, our program constructs a single $k \times k$ matrix $M$, and it does so only once rather than actually tiling it as shown in \eqref{shared_matrix}.
We further found that a more aggressive gradient clipping was needed when training Seq2seq-LSTM models that use VVMAs; otherwise, the gradient grew extremely large and as a result eventually overflowed. We believe that this is because gradients accumulate as we propagate them back to a single small matrix $M$. 

\section{Results} \label{results}

In this section, we present empirical results showing that VVMAs can substitute different types of weight matrices in neural networks (NNs). Specifically, we use our VVMAs in Seq2seq-LSTM and Transformer NMT. We report some theoretical speedups that VVMAs provide when using a TPU-style architecture. We then present a small ablation study where we modify our VVMAs by removing the diagonal terms $D_{i, j}$ or by varying the value of $k$. We also compare VVMA to standard low-rank approximations. Finally, we show that our technique extends to language modelling with Transformer-XL, and beyond NLP tasks.

Unless otherwise noted, all results in this section use VVMAs with $k = 32$. That is, the matrix $W$ in the neural network is replaced with a VVMA that uses a $32 \times 32$ matrix $M$ along with $32 \times 32$ diagonal matrices $D_{i, j}$ as shown in {\eqref{shared_matrix}}.

\subsection{Neural Machine Translation}\label{section:nmt}

We tested our VVMAs on NMT: we integrated them as part of Seq2seq-LSTM and Transformer models, as they are most commonly used today.

\subsubsection{Sequence-to-Sequence Models}
For the Seq2seq-LSTM models \cite{cho2014gru,sutskever2014seq2seq}, we slightly modified the code by {\citet{tf_code}}, and we ran it on the two benchmarks provided in the repository. 
In particular, we used WMT datasets to train German-English and English-German models. We further used IWSLT datasets to train Vietnamese-English and English-Vietnamese models. We prepared the datasets according to the instructions found in the repository. For the German-English and English-German models, we used newstest2015 for testing. 

Both models are Seq2seq models with LSTM layers and attention mechanism. We used four VVMAs for the LSTM cells: for the forget gate, for the input gate, for the output gate, and for the cell state vector. We also used VVMAs for the matrices in the attention mechanism.

For the Seq2seq-LSTM models, we decreased the gradient clipping value from 5 to 1 in order to prevent the gradient from overflowing. We also decreased the batch size to 32, to fit the models on a single GPU. We trained for 340,000 iterations for German-English and English-German, and 48,000 for Vietnamese-English and English-Vietnamese. 

The results comparing the original models with models that use VVMAs are shown in Table~{\ref{table:all:nmt}}. We can see that the BLEU scores decrease when using VVMAs, which should be expected given that the overall number of parameters in the model decreased noticeably. Overall, when taking into account the number of parameters, the observed decrease in the BLEU scores is very reasonable.

\subsubsection{Transformer Models}\label{section:transformer}

For the Transformer models {\cite{transformer}}, we replaced the matrices in the feed-forward layers with VVMAs.\footnote{We modified code from \url{github.com/tensorflow/models/tree/master/official/transformer}} We trained these models on WMT datasets for German-English and English-German translation. We prepared the datasets according to the instructions found in the repository that we modified. 
We used the base Transformer models with a hidden size of 512 (rather than the big models, which have a hidden size of 1024). We trained these models with a batch size of 2048 for 6 epochs. 

In Table~\ref{table:all:nmt}, we present our results on the Transformer models with VVMAs. We achieved reasonable BLEU scores compared to the original Transformer. For German-English, the original model had 61.4M parameters and an uncased test BLEU score of 29.66. The VVMA model had 37M parameters and a BLEU score of 28.5. For English-German, the original model had 61.4M parameters and a BLEU score of 24.57. The VVMA model had 37M parameters and a BLEU score of 23.13. To recap, each matrix in these models was replaced by VVMAs except for the embedding and the projection matrices. We found that restricting these with the VVMA constraints had a sizable negative impact on performance.

\subsection{Theoretical Speedups}\label{section:speedups}

We also calculated two measures for the inference time of the models described in Section~{\ref{section:nmt}}: (\emph{i})~the estimated number of clock cycles, and (\emph{ii})~the number of floating point operations (FLOPs). 
Both roughly correspond to the real time needed to perform the inference at run time.
We computed the former for a TPU-style architecture with one matrix multiply unit of size $k \times k$, and we estimated the latter for the original and the VVMA models using Equations {\ref{original_clocks}} and {\ref{weight_sharing_clocks}} with $k=32$, $t=1$, and sequence lengths of 25. 
Note that the vector-vector operation before $M$ takes zero extra clock cycles, as illustrated in Figures {\ref{figure:1}} and {\ref{figure:2}}. 

This happens because we pipeline these vector-vector processes as we feed the data into the matrix multiply unit. Moreover, we initialize these operations while loading weights into the matrix multiply unit.
We used a TensorFlow profiling tool in order to measure the number of FLOPs in our models. Looking at Table {\ref{table:all:nmt}}, we can see that the original Seq2seq-LSTM models require three to four times more clock cycles and roughly twice as many FLOPs compared to the VVMA models. 

For the Transformer models with VVMAs, we saw less noticeable speed-ups. For similar accuracy, the estimated number of clock cycles and FLOPs were roughly 1.7 and 1.5 times more in the original Transformer models compared to models with VVMAs. This is expected since we use VVMAs only in the feed-forward layers. We tried to use VVMAs for the attention layers as well, but this led to larger decrease in accuracy, due to the significant reduction in the number of parameters.

As the Transformer is already getting noticeable impact in industrial settings, e.g.,~for machine translation and Web search, there is active research in developing more efficient Transformer architectures~\citep{Sanh2019DistilBERTAD, Kitaev2020ReformerTE,Beltagy2020LongformerTL,Zaheer2020BigBT}. Thus, with each new version of a Transformer architecture, new VVMA experiments would be needed in order to measure the potential improvements in efficiency that VVMA would yield.

\subsection{Ablation Study} 

Next, we performed an ablation study for the Seq2seq-LSTM models described in Section {\ref{section:nmt}} for the English-Vietnamese machine translation task. In particular, we slightly modified the VVMAs in the Seq2seq-LSTM models by removing the diagonal terms ${D_{i, j}}$ or by changing the value of $k$. 

Here, we trained with a batch size of 32 for 48,000 steps. In order to prevent the gradient from overflowing, we needed to multiply the shared matrix $M$ by a scaling factor of 0.1 when removing the diagonal terms $D_{i, j}$. 
The results are shown in Table {\ref{table:2}}. We can see that removing the diagonal terms significantly decreases the BLEU scores for our models, while changing the value of $k$ has no significant impact. Additionally, Figure~\ref{figure:3} presents BLEU scores as the number of clock increases. We can see that compared to their original counterparts, VVMA models do not yield degradation in performance when then number of clocks gets large.

\begin{table*}[tb]
  \centering
  \small
  \begin{tabular}{lccccrc} \toprule		
    \bf Architecture & \bf $k$ & \bf Diags & \bf \# Params & \bf BLEU & \bf \# Clocks & \bf FLOPs \\ \midrule
    Original & N/A & N/A & 
    27.5M & 25.34 & 34.8M & 54.9M
    \\
    VVMA & 32 & T &
    17.1M & 24.42 & 10.3M & 34.1M   
    \\
    VVMA & 32 & F & 
    16.7M & 15.62 & 10.3M & 33.8M
    \\
    VVMA & 16 & T &
    17.4M & 24.76 & 22.7M & 34.8M       
    \\
    VVMA & 64 & T &
    16.9M & 23.96 & 5.0M & 33.9M
    \\ \bottomrule
  \end{tabular}
  \caption{Ablation study for English-Vietnamese NMT with Seq2seq-LSTM models. Here, $k$ is the size of $M$ in VVMAs, Diags shows whether diagonal terms are present (T=true, F=false), then follow the number of parameters, BLEU score, number of clocks and FLOPs. Original's clock is on a TPU with a block size of 32.}\label{table:2}
\end{table*}

\begin{table}[tb]
  \small
  \centering
  \begin{tabular}{lrcccc} \toprule		
    \bf Architecture & \multicolumn{1}{c}{\bf $k$}  & \bf \# Params  & \bf \# Clocks & \bf PPL  \\ \midrule
    Original  & N/A & 151.1M  & 99.4M & 24.05  \\
    VVMA & 32  & 138.2M  & 67.0M & 30.70 \\
    VVMA & 64  & 138.1M  & 35.9M & 30.55 \\
    QRNN & N/A & 151.0M  & N/A   & 33.0 \\
    \bottomrule
  \end{tabular}
  \caption{
Language modeling on WikiText-103 using Transformer-XL with and without VVMA, as well as using QRNN. (Original: TPU with a block size of 32.) }\label{table:transformer-xl}
\end{table}

\subsection{Comparison to Standard Low-Rank Approximation}

First, note that the rank of VVMA is maximum $k$ for a $k \times k$ sharing matrix. To prove that, we can represent the matrix in \eqref{shared_matrix} as a product of matrices of maximal rank $k$. Then, we can use the property that $\mathrm{rank}(AB) \leq \min(\mathrm{rank}(A), \mathrm{rank}(B))$.

Second, we compare to low-rank approximation. We fix $k=128$ and we choose $n=$1,024; 2,048; 4,096. We sample a random matrix and we fit VVMA parametrization to it using Adam~\citep{Kingma2015AdamAM} with a learning rate of $0.0001$ ran for $30,000$ steps, and using the Frobenius norm as a loss. We do the same experiment with $UV^\top$ low-rank rank-$p$ approximation, where $p$ is chosen to match the number of parameters in VVMA.  Additionally, we use Eckart--Young--Mirsky’s theorem to get the Optimal low-rank fit.
Table~\ref{tbl:low-rank} shows some Frobenius norm losses from these experiments.
We can see that VVMA's expressiveness is comparable to standard low-rank approximation; note, however, that standard low-rank approximation does not yield the inference speedups of VVMA.

\begin{table}[tb]
  \small
  \centering
  \begin{tabular}{crrr} \toprule		
    \bf $n$ / Fit ($\times 10^3$) & \bf VVMA  & \bf Low-rank & \bf Optimal \\ \midrule
    1,024 & 3.0  & 2.9 & 2.9 \\
    2,048 & 6.1  & 5.9 & 5.8 \\
    4,096 & 12.2 & 11.9 & 11.7
    \\ \bottomrule
  \end{tabular}
  \caption{VVMA's closeness of fit to a target matrix is comparable to that of (\emph{i})~standard low-rank approximation and (\emph{ii})~optimal approximation, but it is orders of magnitude faster at inference time.}\label{tbl:low-rank}
\end{table}

\subsection{Extension to Language Modelling}
Even though the main focus of this paper is the contribution of VVMA to neural machine translation, we also demonstrate that VVMA is compatible to state-of-the-art language modelling architectures. For that purpose, we perform an experiment on WikiText-103~\citep{Merity2017PointerSM} using the Transformer-XL model~\citep{Dai2019TransformerXLAL}. 

In this experiment, we directly integrate VVMA into the Transformer-XL architecture, keeping all hyper-parameter values as in the original Transformer-XL paper \cite{Dai2019TransformerXLAL}, except for reducing the batch size to 30, in order to fit the optimization on two GPUs. We chose to replace the weights of the attention mechanism with VVMA. Replacing the weights of the positional feed-forward layers drastically decreases the number of parameters, which yields poor performance, as Transformer-XL's perplexity is sensitive to the number of parameters. We present our results in Table~\ref{table:transformer-xl}, where we can see that VVMA with a block size of 256 yields reasonable performance, and the perplexity decreases noticeably with the reduction of parameters.

\subsection{Extension to Other Areas}

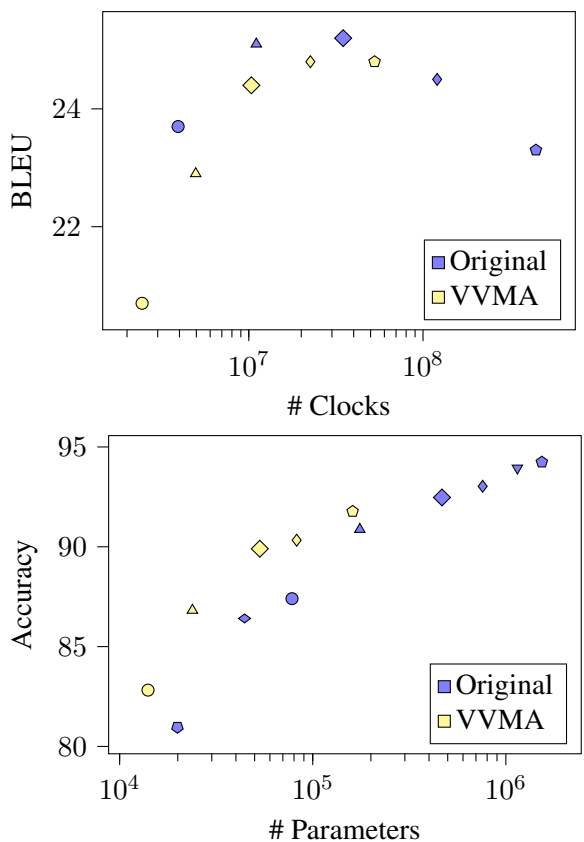
\begin{figure}[tb]
  \begin{tikzpicture}
    \begin{semilogxaxis}[
      width=7.8cm,
      height=5.8cm,
      xlabel=\# Clocks,
      ylabel=BLEU,
      mark size=2.25,
      legend cell align=left,
      legend pos=south east,
      legend entries={Original, VVMA}, 
      tick align=outside,
      every x tick/.style={color=black, thin},
      every y tick/.style={color=black, thin},
      xlabel near ticks,
      ylabel near ticks,
      xtick pos=left,
      ytick pos=left,
      xtick={1000000, 10000000, 100000000},
      scatter/classes={
        original_legend={mark=cube*, fill=blue!50}, 
        weight-sharing_legend={mark=square*, fill=yellow!50}, 
        original2-128={mark=*, fill=blue!50},
        original2-256={mark=triangle*, fill=blue!50},
        original2-512={mark=square*, fill=blue!50, rotate=45},
        original2-1024={mark=diamond*, fill=blue!50},
        original2-2048={mark=pentagon*, fill=blue!50},
        original4-512={mark=triangle*, fill=blue!50, rotate=180},
        weight-sharing2-128={mark=*, fill=yellow!50},
        weight-sharing2-256={mark=triangle*, fill=yellow!50},
        weight-sharing2-512={mark=square*, fill=yellow!50, rotate=45},
        weight-sharing2-1024={mark=diamond*, fill=yellow!50},
        weight-sharing2-2048={mark=pentagon*, fill=yellow!50},
        weight-sharing4-512={mark=triangle*, fill=yellow!50, rotate=180}
        weight-sharing98={mark=pentagon*, fill=yellow!50}
        }
      ]
      \addplot[scatter, only marks, scatter src=explicit symbolic]
      table[meta=label] {
        x y label
        3928550 23.7 original2-128
        11038650 25.1 original2-256
        34803650 25.2 original2-512
        120512850 24.5 original2-1024
        444648050 23.3 original2-2048
        2440550 20.7 weight-sharing2-128
        4957050 22.9 weight-sharing2-256
        10347650 24.4 weight-sharing2-512
        22559250 24.8 weight-sharing2-1024
        52704050 24.8 weight-sharing2-2048
      };
      \legend{Original, VVMA}
    \end{semilogxaxis}
  \end{tikzpicture}
  \begin{tikzpicture}
    \begin{semilogxaxis}[
      width=7.8cm,
      height=5.8cm,
      xlabel=\# Parameters,
      ylabel=Accuracy,
      mark size=2.25,
      legend cell align=left,
      legend pos=south east,
      legend entries={Original, VVMA}, 
      tick align=outside,
      every x tick/.style={color=black, thin},
      every y tick/.style={color=black, thin},
      xlabel near ticks,
      ylabel near ticks,
      xtick pos=left,
      ytick pos=left,
      xtick={10000, 100000, 1000000},
      scatter/classes={
        original_legend={mark=cube*, fill=blue!50}, 
        weight-sharing_legend={mark=square*, fill=yellow!50}, 
        original8={mark=*, fill=blue!50},
        original14={mark=triangle*, fill=blue!50},
        original32={mark=square*, fill=blue!50, rotate=45},
        original50={mark=diamond*, fill=blue!50},
        original98={mark=pentagon*, fill=blue!50},
        original74={mark=triangle*, fill=blue!50, rotate=180},
        original8_c8={mark=pentagon*, fill=blue!50, rotate=180},
        original14_c8={mark=diamond*, fill=blue!50, rotate=90},    
        weight-sharing8={mark=*, fill=yellow!50}, 
        weight-sharing14={mark=triangle*, fill=yellow!50},
        weight-sharing32={mark=square*, fill=yellow!50, rotate=45},
        weight-sharing50={mark=diamond*, fill=yellow!50},
        weight-sharing98={mark=pentagon*, fill=yellow!50}}
      ]
      \addplot[scatter, only marks, scatter src=explicit symbolic]
      table[meta=label] {
        x y label
        466970 92.47 original32
        758618 93.03 original50
        1147482 93.94 original74
        1536346 94.24 original98
        78106 87.4 original8
        175322 90.87 original14
        19922 80.96 original8_c8
        44338 86.41 original14_c8
        14026 82.81999826 weight-sharing8
        23818 86.82000041 weight-sharing14
        53194 89.8999989 weight-sharing32
        82570 90.32999873 weight-sharing50
        160906 91.76999927 weight-sharing98
      };
      \legend{Original, VVMA}
    \end{semilogxaxis}
  \end{tikzpicture}
  \caption{BLEU scores and validation accuracy as a function of the number of trainable parameters in the original and in the VVMA Seq2seq-LSTM models for English-Vietnamese (top) and ResNet~\citep{he2016} models on CIFAR-10 (bottom). The number of parameters is varied by changing the depth and the size of the hidden state. Unique shapes with different colors refer to the same Seq2seq-LSTM model, with the original model in blue and the VVMA model in yellow. 
  }\label{figure:3}
\end{figure}

We further extended our VVMAs beyond NLP, to image classification. We modified the convolutional filters in ResNet \cite{he2016} to use VVMAs and we trained on CIFAR-10.\footnote{We modified code from \url{github.com/tensorflow/models/tree/master/official/resnet}} 
We prepared the CIFAR-10 dataset following the instructions in the repository we modified. 
We trained all ResNet models with a batch size of 128 for 250 epochs. 
Figure {\ref{figure:3}} bottom shows the accuracy of the ResNet models as a function of the number of parameters. We can see that the ResNet models with VVMAs outperform the original ResNet models when keeping the number of parameters fixed.

\section{Discussion} \label{discussion}

Below, we discuss new AI hardware that could optimize inference for neural networks via VVMAs. This hardware would decrease latency at inference time rather than decreasing the training time. 

\paragraph{Tensor Processing Unit.}

As mentioned above, Google's Tensor Processing Units (TPU) has a dedicated matrix multiply unit~\cite{tpu}. We believe that a modified version of the TPU could take advantage of VVMAs.
The necessary modifications would be relatively simple. As illustrated in Figure~{\ref{figure:2}}, we would add a dedicated vector-vector unit before the matrix multiply unit, and we would pipeline it and initialize it at the same time as the matrix multiply unit. As seen in Section~\ref{section:speedups}, this would noticeably decrease the number of inference clock cycles in Seq2seq-LSTM models. 

\paragraph{Tensor Cores.}
NVIDIA's newest GPUs have dedicated matrix multiply units called \emph{Tensor Cores}, which can perform $4 \times 4$ matrix multiplications in a single clock cycle {\cite{tensor_core}}.
Adding vector-vector units before each Tensor Core would make them more efficient for VVMAs.  The largest speedup would come from the time spent loading matrices from the memory into the Tensor Cores. For instance, if multiple Tensor Cores share the same matrix elements, this would decrease the latency when performing inference. 

\paragraph{Optical Processing Unit.}
A newer, more experimental architecture, is to use VVMAs with optical computing. \citet{opu}~proposed to use an Optical Processing Unit (OPU) to perform matrix multiplications at the speed of light. Likewise, it is possible to use an OPU in order to accelerate inference on a neural network.
Note, however, that the OPU would run into some of the same problems that the TPU has. That is, there will be a large delay when loading the matrix weights from the memory onto the OPU. Thus, we propose to add an electronic vector-vector unit before the OPU, which would be pipelined and initialized as weights are loaded onto the OPU.
This extra unit will not increase the overall latency of a system that uses an OPU because the input vectors will still need to be fetched from the digital electronic memory. Likewise, performing vector-vector operations with the input data will not significantly increase the latency of the entire system.

\section{Conclusion and Future Work}
\label{conclusion}

We have proposed a novel vector-vector-matrix architecture for low-latency inference, and we have demonstrated theoretical and empirical speed-ups for Seq2seq-LSTM and Transformer models, with application to neural machine translation, language modeling, and image classification. We hope that this work would bring the novel concept of AI \textit{co-design} (between software and hardware) to the domain of NLP applications.

In future work, we plan to optimize the low-level code and to develop new hardware to deploy VVMAs in real-world applications. Distilling models to their VVMA counterparts would be an interesting experiment, and potentially an orthogonal enhancement to pre-existing frameworks~\citep{Sanh2019DistilBERTAD}. VVMAs could also be an orthogonal contribution to other factorizations of NLP models, such as in~\citep{Lan2020ALBERT}.

\bibliography{anthology,emnlp2020}
\bibliographystyle{acl_natbib}

\appendix

\end{document}